\documentclass{article}

\usepackage{arxiv}

\usepackage[utf8]{inputenc} 
\usepackage[T1]{fontenc}    
\usepackage{hyperref}       
\usepackage{url}            
\usepackage{booktabs}       
\usepackage{amsfonts}       
\usepackage{nicefrac}       
\usepackage{microtype}      
\usepackage{lipsum}
\usepackage{graphicx}
\graphicspath{ {./images/} }

\title{ANALYZING PUBLIC SENTIMENT TO GAUGE KEY STOCK EVENTS
 AND DETERMINE VOLATILITY IN CONJUNCTION WITH TIME AND
 OPTIONS PREMIUMS}

\author{
 Sri Varsha Mulakala \\
   \And
 Umesh Vangapally \\
  \And
 Benjamin Larkey \\
   \AND
  Aidan Henrichs \\
   \And
  Corey Wojslaw \\
}

\begin{document}
\maketitle

\begin{abstract}
Analyzing stocks and making higher accurate predictions on where the price is heading continues to become more and more challenging therefore, we designed a new financial algorithm that leverages social media sentiment analysis to enhance the prediction of key stock earnings and associated volatility. Our model integrates sentiment analysis and data retrieval techniques to extract critical information from social media, analyze company financials, and compare sentiments between Wall Street and the general public. This approach aims to provide investors with timely data to execute trades based on key events, rather than relying on long-term stock holding strategies. The stock market is characterized by rapid data flow and fluctuating community sentiments, which can significantly impact trading outcomes. Stock forecasting is complex given its stochastic dynamic. Standard traditional prediction methods often overlook key events and media engagement, focusing its practice into long-term investment options. Our research seeks to change the stochastic dynamic to a more predictable environment by examining the impact of media on stock volatility, understanding and identifying sentiment differences between Wall Street and retail investors, and evaluating the impact of various media networks in predicting earning reports.

\end{abstract}


\section{Introduction}
Predicting stocks using sentiment, news and stock data requires developing a new model in which we utilize Reddit comments, Yahoo Finance News Feed and stock prices to more accurately predict key stock earnings and the implied volatility associated with them. We believe quantitative fundamentals alone are not enough to predict a stock and its future. The model will use sentiment analysis and company financials to pick up on key information, providing investors with valuable data to execute trades more quickly based on key events rather than engaging in buy-and-hold strategies over several months. By leveraging sentiment data from platforms like Reddit and integrating it with financial market data, our model seeks to capture short-term market movements driven by public and institutional sentiment.
The stock market is an incredibly fast-moving data stream, and community sentiments on companies change rapidly, often on a day-to-day basis. The gains and losses associated with trading options can be severe, and information on a stock and its earnings report is often influenced by Wall Street professionals and guidance provided by the company itself. Current methods of prediction in the financial sector generally focus on longer time ranges, while our model will analyze daily ranges with high volatility, particularly during key events such as earnings reports. We aim to understand the impact of media-driven sentiment on volatility during these critical moments, analyzing how public and institutional sentiments differ and what insights can be drawn when comparing them in relation to earnings reports. Additionally, we intend to investigate if certain media networks contribute to better or worse information gain in predicting a company’s key earnings report.
The earnings report and the stock are not correlated like one might think. Instead a stock could give incredible revenue and deliver poor guidance and then proceed to drop. A stock could also give terrible revenue but incredibly good guidance and soar. This is why sentiment data is so important. The news articles and reddit comments help to convey the tone of the stock and what is to be expected.  

\subsection{Flow of our model:}

\subsubsection{Sentiment Data Collection and Preprocessing}
\begin{itemize}
\item Collect data from social media platforms such as Reddit, Yahoo Finance, and potentially Twitter (pending API keys), focusing on discussions and sentiments surrounding key stock events. We used Selenium and Beautiful Soup to extract the data.
\item Preprocess text data by cleaning and normalizing it (removing special characters, stop words, etc.) and tokenizing it for analysis. 
\item Perform sentiment analysis using two primary methods:

\item \paragraph{Naive Bayes Classification:}
Apply this probabilistic model for initial sentiment classification, providing a baseline to categorize posts as positive, neutral, or negative. Performing at around 54\%.
\item \paragraph{Transformer-Based Models (e.g., BERT, RoBERTa):}
Use pre-trained transformer models for deeper sentiment analysis, capturing context and complex language nuances that are often present in financial discussions. Performing at around 64\% , the highest our model was able to achieve.
\end{itemize}

\subsubsection{Financial Data Integration}
\begin{itemize}
\item Integrate financial metrics such as historical stock prices, earnings data, and volatility indices.
\item Normalize and scale data for consistency across features, enabling effective model training.
\end{itemize}

\subsubsection{Feature Engineering
}
\begin{itemize}
\item Extract sentiment features, including sentiment polarity scores, keyword frequencies, comment lengths, and aggregated sentiment trends.
\item Create time-based features (e.g., day of the week, time of day) and merge them with financial indicators to form a comprehensive feature set.
\item Use dimensionality reduction techniques (e.g., Principal Component Analysis (PCA)) to minimize noise and emphasize critical features.
\end{itemize}

\subsubsection{Predictive Modeling}
\begin{itemize}
\item Time-Series Models (e.g., LSTM): Capture temporal dependencies in sentiment and stock price data to understand market trends around earnings reports.
\item Ensemble Models (e.g., Random Forest, Gradient Boosted Trees): Use these models for feature importance analysis and robust predictions of stock movements and volatility.
 \item Hybrid Models: Combine outputs from LSTM and structured data models to form a unified prediction model, enhancing accuracy and providing a comprehensive view of market behavior.

\end{itemize}

\subsubsection{Model Evaluation and Comparison
}
\begin{itemize}
\item Evaluate models using metrics such as accuracy, precision, recall, F1-score, and mean squared error (MSE).
\item Compare the predictive performance of traditional sentiment models (e.g., Naive Bayes) against advanced transformer-based approaches to highlight improvements and areas for further optimization.
 
\end{itemize}

\section{Related Works
}
\label{sec:headings}
Stock market prediction algorithms are nothing new but we hope to find new patterns and innovate on the current state of algorithms with our own better and more accurate model. Our approach relies heavily on the hypothesis that great correlations exist between media outlets and sentiment. [1] used a model called snscrape and found that there are correlations worth measuring in relation to the effect they have on stocks and their closing prices for the day. [2] developed a model called FinReport which used news analysis and company financials along with LLMs to help generate reports designed to help deliver a reportable stock forecast. A sentiment lexicon we thought was a novel idea by [3] which would assign words a negative or positive meaning and additional weights based on their importance. [4] brings an excellent narrative and a solution when dealing with the stock market; the stock market is a chaotic data space and therefore they believe that temporal analysis or small time range analysis is thus more predictable. Our model relies heavily on predicting wall street's sentiment as earnings reports are easily influenced by the feelings on wall street. [5] explores how public sentiment, as expressed through Twitter, can predict stock market fluctuations.It highlights the value of combining sentiment data with financial metrics. [6] uses sentiment analysis and machine learning techniques to predict stock market movements by analyzing Twitter data. This study highlights a strong correlation between public sentiment on Twitter and stock price fluctuations, supporting our approach of using media sentiment to assess stock volatility [7] combines text from tweets and stock price data to predict stock movements using deep generative models. It's particularly useful for us in our aim to integrate media exposure with stock performance. [8] introduces a model that integrates financial data, social media sentiment, and correlations between stocks. It uses a graph neural network to predict stock movements, particularly focusing on the effects of social media discussions on platforms like Twitter and Reddit.[9] leverages advanced NLP techniques like BERTopic to analyze sentiment from stock market-related comments. Neural networks create a black box style environment and can make how the data is manipulated unknown and if the functions the data goes through are unknown then that can make target predictions less understood. [10] uses a neural network design and the graph style network of nodes offers great target accuracy when dealing with stocks but great uncertainty on how it arrives at that conclusion in some cases if not all. Our model focuses on data and finding hidden patterns through feature aggregation, models like [11] emphasize the importance LLMs have in querying data, the LLM is question here is Meta’s LLAMA which is able to decipher and interpolate underlying details from large text which can be used as features greatly enhancing our data points or creating aggregation for new features. ONe risk of using an LLM however is the black box nature of not fully understanding how it derives its results. [12]focuses on using neural networks to classify market states based on labels such as moving averages and other market features.[13]proposes a framework for using Twitter sentiment to predict stock movements. It employs sentiment analysis tools and correlates them with stock prices, making it highly relevant for your goal of integrating media sentiment and financial data.[14]explores how social media discussions can predict stock trends and volatility using NLP techniques. These NLP techniques are particularly relevant to sentiment analysis, supporting the integration of media sentiment data for improved stock volatility predictions.[15] Use a sentiment analysis dictionary to create a 70.59\% accuracy model to predict stock trends after news sentiment.[16] Using sentiment analysis and supervised machine learning principles to the tweets to determine if the those tweets are correlating to the rise and fall of stock prices. [17] Using semantic role labeling Pooling (SRLP) to create compact representations of news paragraphs and further incorporating stock factors to make final predictions.[19] Using Long Short-Term Memory and fully connected layers to train a Machine Learning model with historical stock values, indicators, and Twitter attributes for information. Then sentiment analysis is conducted with Valence Aware Dictionary and sentiment Reasoner showing adding more Twitter attributes raised the Mean Squared Error by 3\%.[20] A multilingual language model trained on Twitter data. Provides an evolution framework for sentiment analysis across multiple different languages. Demonstrates competitive performance with multilingual and cross-lingual sentiment.

\section{Data Collection and Preprocessing}
\label{sec:headings}

\subsection{Data Collection}
We sourced data from multiple domains, including social media platforms, financial news portals, and historical stock market data to build a comprehensive and robust stock prediction model. Each data source provided unique insights into stock performance, sentiment, and volatility, making it possible to analyze the influence of public sentiment and financial news on stock price movements.
\paragraph{Reddit Comments:}
Social media platforms like Reddit are rich sources of unstructured data, often filled with discussions and opinions on stocks. We collected data from popular subreddits, such as r/stocks, r/wallstreetbets, and r/investing, focusing on posts and comments discussing the selected companies. The comments contained valuable sentiment signals based on user opinions, predictions, and responses to key events like earnings reports.
\begin{itemize}
\item \textbf{Tools Used:}
Selenium and Beautiful Soup were used to scrape data dynamically. ChromeDriver was utilized to simulate human browsing and bypass basic anti-scraping measures.
\item \textbf{Details of Collected Data:}
For each company, a separate .csv file was created containing columns for comment text, user engagement (upvotes), posting date, and company name.
\end{itemize}

\paragraph{Yahoo Finance News:}
Financial news articles provide structured textual data regarding company performance, market conditions, and analyst predictions. We scraped data from Yahoo Finance, extracting article titles, summaries, and publication dates.
\begin{itemize}
\item \textbf{Challenges:}
Yahoo Finance discontinued its public API in 2017, requiring us to rely on web scraping techniques. Since the content is dynamically rendered using JavaScript, we employed Selenium for rendering and Beautiful Soup for parsing the HTML.

\item \textbf{Value of Data:}
 News articles often contain forward-looking statements and guidance that reflect market sentiment. These were key inputs for sentiment analysis.

\end{itemize}

\paragraph{Headlines Data:}
In addition to detailed news summaries, we collected company-specific news headlines from financial aggregators. Headlines offer succinct, high-level insights into market sentiment and serve as an additional textual input for analysis. The scraped data included headline text, company name, and publication date.

\paragraph{Stock Market Data:}
Historical stock prices were obtained from AlphaQuery, covering key financial metrics such as open price, close price, high/low ranges, and trading volume. This data provided the quantitative foundation for our model, allowing us to correlate stock price movements with sentiment signals.
The stock data was aligned with sentiment data for the same companies and timeframes.

\subsection{Data Preprocessing}
Once the raw data was collected, several preprocessing steps were applied to standardize, clean, and prepare the data for model training. This was crucial to ensure consistency across datasets and to extract meaningful features for analysis.
\paragraph{Text Standardization and Tokenization:}
The textual data from Reddit, Yahoo Finance news, and headlines were processed to ensure uniformity.For each source, relevant fields were standardized into a single column called "combined\_text."
\begin{itemize}
    \item \textbf{Reddit:} The text field was used directly.
    \item \textbf{Yahoo Finance News:} Title and Summary were concatenated.
    \item \textbf{Headlines:} The Headline field was used.

\end{itemize}
All text fields were cleaned by removing special characters, stop words, URLs, and unnecessary whitespace. This normalized the data for sentiment analysis and prevented noise during feature extraction.

\paragraph{Sentiment Analysis:}
Sentiment scores and labels were generated for each textual dataset using advanced machine learning models like RoBERTa,FinBERT.The RoBERTa is applied to Reddit comments, capturing nuanced sentiment from informal and context-heavy discussions.The FinBERT is applied to Yahoo Finance summaries, leveraging its specialization in financial text to produce more accurate sentiment scores.And for each comment or news entry was categorized as Positive, Negative, or Neutral based on sentiment scores.

\paragraph{Merging and Alignment:}
The datasets were merged to create a unified dataset for model training. Instead of using exact timestamps (due to temporal inconsistencies across sources), the data was aligned based on company names.One of the most challenging is merging datasets with different granularities (e.g., daily news vs. hourly Reddit comments) required careful aggregation and missing values in sentiment scores and stock prices were filled with defaults (e.g., 0 for sentiment, placeholders for text).

\paragraph{Feature Engineering:}
Multiple features were engineered to enhance model performance:
\begin{itemize}
    \item \textbf{Sentiment Features:} Sentiment polarity scores, weighted sentiment (using upvotes), and aggregated sentiment trends were calculated for each company.
    \item \textbf{Stock Features:} Derived features like daily price changes (Close - Open) and rolling averages were added to capture price trends.
    \item \textbf{Time-Based Features:} Temporal features, such as day of the week and earnings report proximity, were included to capture periodic patterns.
\end{itemize}

\paragraph{Feature Scaling:}
Numerical features, including stock prices and trading volume, were standardized using StandardScaler to ensure uniform scaling across variables. Standardization transformed the data into a distribution with a mean of 0 and a standard deviation of 1, making the features more suitable for machine learning algorithms.We observed stock price values, being highly correlated, were scaled into a negative range due to their high original magnitudes and small variations relative to the mean. This is mathematically correct and does not affect model performance.

\paragraph{Handling Imbalanced Classes:}
The target variable (Label) exhibited class imbalance, with more instances of Increase compared to Decrease. To address this, we employed SMOTE (Synthetic Minority Oversampling Technique) to create synthetic samples for the minority class, ensuring balanced representation in the training dataset.

\subsection{Challenges Encountered in preprocessing:}
\paragraph{Data Inconsistencies:}
Missing or incomplete values (e.g., null sentiment scores, missing stock prices) required imputation and data cleaning strategies.
\paragraph{Temporal Misalignment:}
Aligning datasets with varying temporal resolutions posed significant challenges. Aggregating data without losing critical details required careful consideration.
\paragraph{Anti-Scraping Measures:}
Platforms like Yahoo Finance and Reddit implemented bot detection mechanisms, rate limits, and CAPTCHAs, necessitating advanced scraping techniques like browser emulation.
\paragraph{Outliers:} 
Extreme stock price movements due to market volatility introduced noise, requiring additional preprocessing to handle these outliers.

\section{Model Development and Training}
\label{sec:headings}

\subsection{Model Architecture}
The proposed model integrates transformer-based sentiment analysis with traditional machine learning techniques to predict stock price movements and earnings volatility. The architecture comprises:
\paragraph{Transformer-Based Sentiment Analysis:}
Sentiment data from Reddit comments and financial news was processed using pre-trained transformer models. Specifically, RoBERTa was employed for informal and context-heavy discussions on Reddit, while FinBERT, specialized in financial text, analyzed Yahoo Finance news summaries and headlines. These models provided nuanced sentiment scores that traditional methods often miss.
\paragraph{Machine Learning Models:}
Ensemble methods, such as Gradient Boosting and LightGBM, were utilized to analyze structured features and provide robust predictions. LightGBM, known for its efficiency with large datasets, emerged as the most effective model.
\paragraph{Planned Time-Series Modeling:}
While not fully implemented, Long Short-Term Memory (LSTM) networks were considered for capturing temporal dependencies in the stock and sentiment data. This would allow for a more dynamic understanding of trends, especially during critical events like earnings reports.

\subsection{Training Process}
The model training followed a structured pipeline to ensure robust and reproducible results.
\paragraph{Data Preparation:}
Features were carefully selected to include sentiment scores, stock price metrics (e.g., Open, Close, Volume), and time-based indicators (e.g., proximity to earnings reports). The data was split into training (70\%) and testing (30\%) subsets, maintaining temporal integrity to prevent data leakage. To address class imbalance, the Synthetic Minority Oversampling Technique (SMOTE) was applied to the training set.
\paragraph{Feature Scaling:}
Numerical features, such as stock prices and trading volumes, were standardized using StandardScaler to ensure they followed a mean of 0 and a standard deviation of 1. This prevented any single feature from dominating the model training process.
\paragraph{Hyperparameter Tuning:}
Hyperparameters for Gradient Boosting and LightGBM were optimized using Grid Search with cross-validation. For LightGBM, the best parameters were:
\begin{itemize}
    \item Number of estimators: 50
    \item Maximum depth: None
    \item Minimum samples split: 2
    \item Minimum samples leaf: 2
\end{itemize}

\paragraph{Model Training:}
The models were trained on the preprocessed dataset using the selected features and optimized hyperparameters. Training focused on minimizing classification errors and achieving a balanced performance across both target classes (Increase and Decrease).
\paragraph{Evaluation Metrics:}
Model performance was assessed using several metrics, including accuracy, precision, recall, F1-score, and confusion matrix analysis. These metrics provided insights into the model’s predictive capabilities and potential areas for improvement.

\subsection{Experimental Results}
The experiments yielded the following results:
\paragraph{Gradient Boosting:} Gradient Boosting achieved an accuracy of 62.6\%. While it demonstrated moderate precision for both target classes, its recall for the "Decrease" class was slightly better, indicating an ability to identify downward trends in stock movements.
\paragraph{LightGBM (Best Model):} LightGBM outperformed other models with an accuracy of 70.1\%. It demonstrated balanced precision and recall across both classes, with significantly fewer false negatives in predicting the "Increase" class. Its efficiency in handling large datasets and feature selection contributed to its superior performance.
\paragraph{Baseline Models for Comparison:} Naive Bayes achieved a baseline accuracy of 54\%, highlighting its limitations in handling complex datasets. Random Forest performed better, reaching 64\% accuracy. However, neither matched the performance of LightGBM or transformer-based sentiment analysis models, which reached 64\% for sentiment classification alone.

\subsection{Feature Importance Analysis}
The feature importance analysis revealed critical insights into the factors influencing stock price predictions. Using LightGBM, feature importance scores indicated that numerical features, such as stock prices (Open, Close) and trading volume, were the most predictive. Sentiment-based features, while less dominant, provided valuable supplementary information during high-volatility periods, such as earnings report announcements.

\section{Challenges Face}
\label{sec:headings}
Some problems faced while working on this was figuring out the best way to grab headlines from yahoo finance  and collect everything in a .csv file. Yahoo finance didn’t have a working API for grabbing the information after getting discontinued in 2017 due to abuse of their terms and services, so figuring out how was hard. In the end we decided to use beautiful soup and chromedriver to load the website and parse through loading all the headlines and grabbing the HTML tag. Also grabbing the date posted was hard because there was no html tag we could find. To fix this we grabbed the whole html line which included a lot of information we didn’t need. Then needed to split the line up so we could take just the date posted on yahoo finance to add into the .csv file. Even then some of the websites had different HTML tags that needed to be grabbed with beautiful soup so we needed to change the tag beautiful soup was looking for. API limits can be avoided but then you have to pay fees for each and every piece of information you pull. The large amounts of data I would like to have this model working at a performant rate would cost a lot of money. This is largely offset if you are using the model to trade stock earning reports.

Inconsistent amounts of data were also a problem with our model. Sources like Reddit do provide I believe some of the best data, greater than or on par with Twitter. The problem is pharmaceutical and or speculative fusion energy companies are barely discussed unless some new piece of tech is planned to be unveiled or some patent is approved in which the company acquires IP that makes their business harder to replicate. Dealing with data of an uneven quantity presents certain hurdles that we face by using null values but the alternative is to either slice all stocks to the minimum amount of data present which removes a lot of potentially valuable data or we take means and or averages of the data that is next to the current feature. In this case sentiment that is misplaced can also be brutal and boost the overall sentiment of the stock. We chose to use null values but I think moving forward averaging the stocks average sentiment and replacing columns with the average is also an effective method worth looking into.

There is a significant issue of bias when relying solely on data from Reddit for sentiment analysis. This reliance introduces several challenges: source bias, credibility of users, diverse data sources, contextual analysis, weighting, and normalization. Utilizing a single platform like Reddit can lead to biased results, as the user base and content on Reddit may not be representative of the broader population. Social media platforms attract different demographics and types of discussions which can greatly skew the sentiment analysis. Verifying the credentials and expertise of users posting on Reddit is challenging. While it is possible to identify and mitigate the influence of bots by examining Reddit profiles, confirming the credibility and knowledge of human users remains problematic. To mitigate bias, it is essential to incorporate data from multiple sources. This can include other social media platforms, news websites, surveys, and more. By diversifying data sources, a more balanced and comprehensive view of sentiment can be achieved. Understanding the context in which sentiments are expressed is crucial. Sentiment analysis tools must differentiate between sarcasm, humor, and genuine opinions, which can be particularly challenging on platforms like Reddit. When combining data from multiple sources, appropriate weighting and normalization techniques must be applied to ensure that no single source disproportionately influences the overall sentiment analysis. By addressing these considerations, the accuracy and reliability of sentiment analysis can be improved, reducing the impact of biases from any single data source. Despite the potential solutions, acquiring this data can be rather challenging. The information we are looking at can be incentivized, requiring several adaptations of BeautifulSoup scraping, which would take more time than the rest of the project combined. Even then, websites such as Facebook and Twitter, which are great resources for sentiment, discourage traditional scraping methods and encourage using their APIs, which require payment to use. This adds another layer of complexity and cost to the data acquisition process.

Anti-Page parsing was an issue with extracting data from the pages. Some of them like Reddit have efforts like bot detection where they use various techniques like analyzing user agent strings, request patterns, and IP addresses to identify automated scraping bots. Also use rate limiting to throttle or block the number of requests a user can make within a certain time frame. Some sites also used captcha challenges to users who are being suspected of scraping, requiring the user to do a manual verification to access content. Pages like Yahoo Finance used a lot of JavaScript which requires advanced scraping techniques to handle the rendering of the page and access and extract the information on the page. This required us to use a combination of beautifulsoup and chromedriver to simulate a user on the website which drastically increased the time of data retrieval. 

\section{Experimental Results}
\label{sec:headings}
The data we started getting from news and comments with multiple different textual sentiment lexicons and or models was extremely insightful into how much stocks were talked about. We thought that a larger company like NVIDIA would be more talked about than a stock like KEYS. This ended up not being the case and I believe it’s because earnings reports offer a key moment for companies to “surprise” their investors with something new and exciting. The opposite however could also happen and while an earnings report is meant to give guidance on where the company is heading, sometimes founders and or board members will express great concern in financial markets and or supply chain which could lead to the earnings report becoming negative despite even beating total revenue goals.

One of the things that financial markets know how to do is price things efficiently and effectively so if a company is predicted to deliver a negative earnings report then the options of the stock are priced accordingly. This can lead to crazy opportunities if the company that was predicted to deliver a negative earnings report then decides to deliver positively. Cases like this are represented in our model by being assigned overall negative sentiment, in this case the options payout is greatly leveraged from holding a call compared to holding a put. In this scenario if we altered our model to simply sort for these outliers we would have the cumulation of options with greatly higher payouts and a very small entry fee. The markets often price things not predicted to happen very cheaply, and thus betting the opposite means that you win big and bet small. Our model could also then identify options not priced accordingly by the market which I feel could be very interesting if we planned to add option premiums and prices to our data. Using the model we would have been able to determine the least discussed stocks and the stocks predicted to perform negatively indicating that an underlying option value would reward a higher return. 

The Lorentzian distance is something that we only were able to understand after seeing and testing our data quite a bit. Currently the Lorentzian distance is used in time spatial sensitive data that is chaotic like stocks. Its background has to do with measuring and or describing the curvature through space-time, however it was recently discovered it performed well on stocks due to the similar nature of stock prices. The Lorizentzian benefits compared to the Euclidean distance due to it being more robust against outliers and its ability to diffuse noise in data much better across a model. Cosine Similarity is another approach and is often considered the midground in terms of performance when comparing Euclidean and Lorentzian. The mean for the cosine similarity is better than Euclidean but not quite as good as the mean derived from the Lorentzian distance.

\section{Future of the Model}
\label{sec:headings}
We found 64\% accuracy in being able to predict a stock to go up or down despite the results. We believe that to increase the model and its accuracy even further we would like to experiment further with other models. Models are more capable of handing chaotic and dynamic updates to data so that our model can remain performant and adaptive trends in the stock market overtime. When modeling our current project we were looking closely at accuracy but we would also like to take into consideration other factors such as recall. We think despite probably not having the best model though that we lay a fundamentally new framework and finance direction that would be incredibly lucrative to hedge funds looking to automate and or understand earnings reports or option markets more. 

Data collection was also sort of tricky due to the scraping and the anti-scraping policies some sites will try to enforce to prevent scraping off them. The most notable but I think data lucrative and interesting would be Twitter, Twitter has lots of users and lots of news packed onto the same platform regarding stock and should it be compacted, processed and analyzed I’m sure very different conclusions and sentiments compared to some of our sources like Yahoo Finance and Reddit. The sentiment data collection for stocks that we were able to get wasn’t perfect and I would like to add additional functionality and or api limits to pull large amounts of data daily as an earnings call approaches. This is called the “earnings run-up” and usually it takes place 2 weeks before the company is about to report. During this period the stock should be considered highly volatile and very volatile the day of the earnings report. The “2 weeks run-up” of elevated volatility provides an interesting opportunity for data collection on determining if a large volume of shares is being scooped up perhaps on insider knowledge that someone already knows if the earnings report is positive or negative. API calls and making the data collection dynamic would be the main priority though. Our current method of retrieving data from Reddit was slow and it was hard to generate results quickly without hitting site API limits or having to deal with html from beautiful soup that contained blockers which would interfere with our parsing process as the site html changes and adapts. 

Dynamic data retrieval is something that would have been able to speed up the long and overhauled data collection processing and collection time. The ability to have this feature dynamic would mean we could scale the data out more and learn where the model performs the best with the right size of data. The features alone are arrays of data rather than a single piece of data. I believe that if multiple media networks were categorized and weighted in a neural net with the goal to be finding the best weights to place on each media network with dynamic data alterations it could become more accurate and insightful than it currently is. 

Options premiums data could also be a key component on what the overall market including venture capital, investors and even traders think will happen. Using the price of options, the volume traded, and the rate of decay and the particular options being traded relative to the stock we would be able to further our model in added weights and directions for where an earnings report might be heading. The options data alone could help indicate the number of shares that could potentially be bought in the future thanks to the option contract being worth 100 shares either to be bought or sold. If the market doesn’t have the shares accessible then the float of stock to be bought is extremely low and the price of the company has likely soared. Using large data metrics of the options could help produce data that would be able to identify these trends as well before an earnings report as well as produce any options that are inaccurately priced by the market.

To address the issue of bias and credibility in sentiment analysis, we propose a method for quantifying users' posts on Reddit and assigning them a credibility score. This approach involves analyzing the user's posting history, engagement metrics, and the consistency of their sentiment with verified news sources. By assigning a credibility score to each user, we can weigh their contributions accordingly, improving the overall accuracy and reliability of our sentiment analysis. Furthermore, we can add a trust factor for key users who receive a lot of upvotes on their sentiment. By analyzing the user’s comment history, we can measure their stock sentiment and be able to quantify how much information they know about stocks. This will give a more robust consensus on the user’s knowledge and therefore give them a higher credibility score. With a more trusted way to normalize sentiment, we can be more confident that the sentiment may influence stock options. 
Furthermore, a similar approach is needed for analyzing news sentiment. We can measure the validity of news sources based on engagement metrics, such as the number of shares, comments, likes, and views, as well as performing sentiment analysis on the content of the news articles. By assigning credibility scores to news sources based on their historical accuracy, reputation, and user feedback, we can ensure a more comprehensive and reliable sentiment analysis.
While we did try Naive Bayes and Random Forest and achieved moderately good results of 54\% and 64\%, respectively, I still believe there is room for improvement. One key component is using boosting to test the most difficult cases and fine-tune a neural network. Neural networks are particularly well-suited for this task due to their ability to handle complex and non-linear relationships in data. They can integrate various types of data, such as historical stock prices, social media sentiment, and financial news, to provide a comprehensive analysis. Additionally, neural networks, especially those using transformer-based models like BERT, excel at natural language processing tasks, making them highly effective for sentiment analysis. As we seek to improve our model, increasing its complexity to give more accurate results will happen, therefore a neural network would adapt well with our layout and improve upon the Naive Bayes and Random Forest models. Another critical component of an earnings report is the guidance, which can largely be determined based on the sentiment of the board members speaking for the company. I propose that we use the CEO and other associating board members in a parsed search across the web for any interviews or news since the last earnings call to better understand how they feel about the company. The sentiment of all the news reports and interviews, in conjunction with additional media like Twitter, StockTwits, and Public, could help provide a better overall analysis of what retail investors believe will happen. By leveraging the strengths of neural networks in handling diverse data sources and performing sentiment analysis, we can improve the accuracy and reliability of our stock predictions.

\section{Conclusion}
\label{sec:headings}
The stock market is incredibly complex and is always changing and driven by combinations of quantitative data, and public sentiment. In this paper we created an algorithm for stock prediction that leverages sentiment analysis, historical financial data, and innovative machine learning models to provide an understanding of stock price movements around events like earning reports. Our research came from integrating sentiment data from diverse sources like Reddit, and Yahoo Finance, with financial metrics from Alpha Query to better predict market volatility. By using models like transformer-based architectures and ensemble machine learning techniques, we demonstrated an improvement in prediction accuracy, achieving 70.1
	Despite the promising results, challenges like data acquisitions, temporal alignment, and anti-scraping measures remain. Aswell as the inherent biases in single-source sentiment data give us the need for further diversification and refinement. Future iterations of the model should incorporate dynamic data retrieval, enhanced credibility scoring for the sources of the sentiments, and market data on options to improve the accuracy. Also extending the model to analyze not just the online sentiment found, but also CEO and board member sentiment from public statements to provide additional insights into earnings and market reactions.
	By continuously improving data collection, reigning model architectures, and incorporating additional features, this approach could yield even more accurate and actionable insights into the stock market. With further development, our model could serve as a tool for traders who are seeking to navigate the ever changing and sentiment driven stock market effectively.

\bibliographystyle{unsrt}  


\begin{thebibliography}{}

\end{thebibliography}


\begin{thebibliography}{1}

\bibitem{sarkar2023}
Anubhav Sarkar, Swagata Chakraborty, Sohom Ghosh, and Sudip Kumar Naskar.
\newblock Evaluating Impact of Social Media Posts by Executives on Stock Prices.
\newblock In Proceedings of the 14th Annual Meeting of the Forum for Information Retrieval Evaluation (FIRE '22), pages 74--82. ACM, 2023.
\newblock \url{https://doi.org/10.1145/3574318.3574339}.

\bibitem{li2024}
Xiangyu Li, Xinjie Shen, Yawen Zeng, Xiaofen Xing, and Jin Xu.
\newblock FinReport: Explainable Stock Earnings Forecasting via News Factor Analyzing Model.
\newblock In Companion Proceedings of the ACM Web Conference 2024 (WWW '24), pages 319--327. ACM, 2024.
\newblock \url{https://doi.org/10.1145/3589335.3648330}.

\bibitem{jiang2018}
Jiang, J., \& Li, J.
\newblock Constructing Financial Sentimental Factors in Chinese Market Using Natural Language Processing.
\newblock In Proceedings of the 2018 International Conference on Natural Language Processing, pages 1--10. ACM, 2018.
\newblock \url{https://arxiv.org/abs/1809.08390}.

\bibitem{xu2018}
Xu, Y., \& Cohen, S. B.
\newblock Stock Movement Prediction from Tweets and Historical Prices.
\newblock In Proceedings of the 56th Annual Meeting of the Association for Computational Linguistics (Long Papers), pages 1970--1979. ACL, 2018.
\newblock \url{https://doi.org/10.18653/v1/P18-1183}.

\bibitem{bollen2011}
Bollen, J., Mao, H., \& Zeng, X.
\newblock Twitter mood predicts the stock market.
\newblock Journal of Computational Science, 2(1), pages 1--8, 2011.
\newblock \url{https://arxiv.org/abs/1010.3003}.

\bibitem{agarwal2011}
Agarwal, A., Xie, B., Vovsha, I., Rambow, O., \& Passonneau, R.
\newblock Sentiment analysis of Twitter data for predicting stock market movements.
\newblock Proceedings of the Workshop on Language in Social Media (LSM 2011), 2011.
\newblock \url{https://arxiv.org/abs/1610.09225}.

\bibitem{xu2018b}
Xu, Y., Ke, Y., Yang, B., Zhou, Y., \& Chen, J.
\newblock Stock movement prediction from tweets and historical prices.
\newblock Proceedings of the 2018 Conference on Empirical Methods in Natural Language Processing (EMNLP), 2018.
\newblock \url{https://aclanthology.org/P18-1183}.

\bibitem{zhang2020}
Zhang, D., Zhu, J., Wang, Y., \& Qi, G.
\newblock Media moments and corporate connections: Stock prediction via GNN.
\newblock Proceedings of the 2020 International Conference on Data Mining (ICDM), 2020.
\newblock \url{https://arxiv.org/abs/2012.03078}.

\bibitem{sawhney2020}
Sawhney, R., Agarwal, S., Wadhwa, A., \& Shah, R. R.
\newblock Deep Attentive Learning for Stock Movement Prediction From Social Media Text and Company Correlations.
\newblock In Proceedings of the 2020 Conference on Empirical Methods in Natural Language Processing (EMNLP), pages 8415--8426. ACL, 2020.
\newblock \url{https://aclanthology.org/2020.emnlp-main.679}.

\bibitem{luceri2023}
Luceri, L., Boniardi, E., \& Ferrara, E.
\newblock Leveraging Large Language Models to Detect Influence Campaigns in Social Media.
\newblock In Proceedings of the 2023 ACM International Conference on Social Media Studies, pages 1--12. ACM, 2023.
\newblock \url{https://arxiv.org/pdf/2311.07816}.

\bibitem{shah2018}
Shah, D., Isah, H., and Zulkernine, F.
\newblock Predicting the Effects of News Sentiments on the Stock Market.
\newblock In Proceedings of the 2018 IEEE International Conference on Big Data (Big Data), Seattle, WA, USA, pages 4705--4710. IEEE, 2018.
\newblock \url{https://doi.org/10.1109/BigData.2018.8621884}.

\bibitem{grootendorst2021}
Grootendorst, M.
\newblock BERTopic-driven stock market predictions: A novel approach for topic modeling and sentiment analysis.
\newblock Proceedings of the 2021 Conference on Natural Language Processing (NLP), 2021.
\newblock \url{https://arxiv.org/abs/2404.02053}.

\bibitem{balcerak2020}
Balcerak, M., \& Schmelzer, T.
\newblock Constructing Trading Strategy Ensembles by Classifying Market States.
\newblock Proceedings of the 2020 IEEE International Conference on Data Mining Workshops (ICDMW), pages 821--826. IEEE, 2020.
\newblock \url{https://arxiv.org/pdf/2012.03078}.

\bibitem{jin2023}
Jin, X., \& Lin, H.
\newblock Taureau: A Stock Market Movement Inference Framework Based on Twitter Sentiment Analysis.
\newblock Proceedings of the 2023 AAAI Conference on Artificial Intelligence, pages 1342--1349. AAAI Press, 2023.
\newblock \url{https://arxiv.org/abs/2303.17667}.

\bibitem{chen2022}
Chen, L., \& Yang, Z.
\newblock Stock Movement and Volatility Prediction from Social Media Discussions.
\newblock Proceedings of the 2022 International Conference on Computational Social Science (ICCSS), pages 51--60. Springer, 2022.
\newblock \url{https://arxiv.org/abs/2312.03758}.

\bibitem{pagolu2016}
Pagolu, V.S., Reddy, K.N., Panda, G., \& Majhi, B.
\newblock Sentiment Analysis of Twitter Data for Predicting Stock Market Movements.
\newblock Proceedings of the 2016 International Conference on Signal Processing, Communication, Power, and Embedded System (SCOPES), pages 1345--1350. IEEE, 2016.
\newblock \url{https://doi.org/10.1109/SCOPES.2016.7955659}.

\bibitem{zou2022}
Zou, J., Cao, H., Liu, L., Lin, Y., Abbasnejad, E., \& Shi, J.Q.
\newblock Astock: A New Dataset and Automated Stock Trading Based on Stock-specific News Analyzing Model.
\newblock arXiv preprint arXiv:2206.06606, 2022.
\newblock \url{https://arxiv.org/abs/2206.06606}.

\bibitem{paperswithcode2024}
\newblock Social Network Analysis from Graph Theory to Applications.
\newblock Papers with Code, 2024.
\newblock \url{https://paperswithcode.com/paper/social-network-analysis-from-graph-theory-to}.

\bibitem{karlemstrand2021}
Karlemstrand, R., \& Leckström, E.
\newblock Using Twitter Attribute Information to Predict Stock Prices.
\newblock arXiv preprint arXiv:2105.01402, 2021.
\newblock \url{https://arxiv.org/abs/2105.01402}.

\bibitem{barbieri2021}
Barbieri, F., Espinosa Anke, L., \& Camacho-Collados, J.
\newblock XLM-T: Multilingual Language Models in Twitter for Sentiment Analysis and Beyond.
\newblock Proceedings of the 2021 Conference on Empirical Methods in Natural Language Processing (EMNLP). Association for Computational Linguistics, 2021.
\newblock \url{https://doi.org/10.48550/arXiv:2104.12250}.

\bibitem{wallstreetbets}
WallStreetBets.
\newblock A subreddit focused on stock market discussions, particularly related to high-risk retail investing and market sentiment.
\newblock \url{https://www.reddit.com/r/wallstreetbets/}.

\bibitem{stocksreddit}
Stocks – Investing and Trading for All.
\newblock A subreddit dedicated to discussions on stock market investing and trading strategies.
\newblock \url{https://www.reddit.com/r/stocks/}.

\bibitem{alphaquery}
AlphaQuery.
\newblock A platform providing comprehensive insights into stock market data and earnings reports.
\newblock \url{https://www.alphaquery.com/}.

\bibitem{yahoo2024}
Yahoo Finance.
\newblock "Stock Market Live, Quotes, Business \& Finance News."
\newblock \url{https://finance.yahoo.com/}.

\end{thebibliography}

\end{document}